# A Grammar Formalism and Cross-Serial Dependencies[*]

Tore Burheim [†]

**Abstract**

First we define a unification grammar formalism called the Tree Homomorphic Feature Structure Grammar. It is based on Lexical Functional Grammar (LFG), but has a strong restriction on the syntax of the equations. We then show that this grammar formalism defines a full abstract family of languages, and that it is capable of describing cross-serial dependencies of the type found in Swiss German.

## 1 Introduction

Due to their combination of simplicity and flexibility unification grammars have become widely used in computational linguistics in the last fifteen yeas. But this flexibility results in a very powerful formalism. As a result of this power, the membership problem for unification grammars in their most general form is undecidable. Therefore most such grammars have restrictions to make them decidable, e.g. the *off-line parsability constraint* in LFG [KB82]. Even so, the membership problem is NP-complete [BBR87] or harder for most unification grammar formalisms. It is therefore interesting to study further restrictions on unification grammars. Most such studies have been concerned with making the formalism decidable [Joh91, Joh94]. But there has also been work on formalisms for which the membership problem can be decided in polynomial time. GPSG [GKPS85] which was one of the first unification grammar formalisms has only a finite number of possible feature structures and describes the class of context-free languages. Then it follows that we can decide in polynomial time if a given string is a member of the language generated by a GPSG-grammar. In their work, Keller and Weir [KW95] define a grammar formalism with feature structures for which the membership problem can be decided in polynomial time. Here there is no common feature structure for the sentence as a whole, only feature structures annotated to the nodes in a phrase structure tree, with only limited possibilities to share information. In this paper we will study a formalism that lies somewhere in between the most powerful formalisms and the most limited ones. The grammar formalism that we define is based on work by Colban [Col91].

What we here call unification grammars are also called attribute-value grammars, feature-structure grammars and constraint-based grammars. We may divide them in two major groups, those based on a phrase-structure backbone such as LFG and PATR, and those

---

entirely described using feature structures such as HPSG [PS94]. We will here use a context-free phrase structure backbone and add equations to the nodes in the phrase-structure tree as in LFG. These equations will describe feature structures. Due to a restriction that we will impose on the equations in the grammar, the feature structures will be trees that are homomorphic with the phrase structure tree. This homomorphism is interesting from a computational point of view.

## 2 Feature structures

One of the main characteristics of unification grammars is that they are information based. This information is inductively collected from the sentences sub-strings, sub-sub-strings and so on. We will use feature structures to represent this information. There are many ways of viewing, defining and describing feature structures, e.g. as directed acyclic graphs [Shi86], as finite deterministic automata [KR90], as models for first order logic [Smo88, Smo92, Joh88], or as Kripke frames for modal logic [Bla94]. Here we use a slightly modified version of Kasper and Rounds [KR90] definition of feature structures, and we will later use a subset of the equations schemata used by LFG to describe these feature structures. As a basis we assume two predefined sets, one of attribute symbols and one of value symbols. In a linguistic framework these sets will typically include things like *subject, object, number, case* etc. as attribute symbols, and *singular, plural, dative, accusative* etc. as value symbols.

**Definition 1** *A feature structure $M$ over the set of attribute symbols $A$ and value symbols $V$ is a 4-tuple $\langle Q, f_D, \delta_0, \alpha \rangle$ where*

- *$Q$ is a finite set of* nodes,

- *$f_D : D \to Q$ is a function, called the* name mapping function, *where $D$ is a finite set of names,*

- *$\delta_0 : Q \times A \to Q$ is a partial function, called the* transition function,

- *$\alpha : Q \to V$ is a partial function called the* atomic value function.

*We extend the transition function $\delta_0$ to be a partial function $\delta : (Q \times A^*) \to Q$ as follows: 1) For every $q \in Q$, $\delta(q, \varepsilon) = q$, 2) if $\delta(q_1, w) = q_2$ and $\delta_0(q_2, a) = q_3$ then $\delta(q_1, wa) = q_3$ for every $q_1, q_2, q_3 \in Q$, $w \in A^*$ and $a \in A$.*

*A feature structure is* well defined *if it is*

- atomic: *For all $q \in Q$, if $\alpha(q)$ is defined, then $\delta_0(q, a)$ is not defined for any $a \in A$.*

- acyclic: *For all $q \in Q$, $\delta(q, w) = q$ if and only if $w = \varepsilon$.*

- describable: *For all $q \in Q$ there exists an $x \in D$ and a $w \in A^*$ such that $\delta(f_D(x), w) = q$*

*All feature structures are required to be well defined.*

We may also view this as a directed acyclic graph where all the edges are labeled with attribute symbols and some nodes without out-edges have assigned value symbols. In addition we name some nodes, such that each node may have more than one name. We will draw feature structures as graphs; an example is shown in Figure 2.

Some definitions of feature structures require that they have an initial node from which one can reach every other node with the extended transition function. We prefer to use the name mapping function and to require feature structures to be describable. If we instead of the name mapping function add an initial node $q_0$, and replace the name mapping function by a definition of $\delta(q_0, x) = f_D(x)$ for all $x$ such that $f_D(x)$ is defined, we get a feature structure with an initial node as specified from a describable one with names. We will in the rest of the paper view the domain of names as implicitly defined in the name mapping function $f$ and drop $D$ as subscript.

We use equations to describe feature structures, such that a set of equations describes the least feature structure that satisfies all the equations in the set: A feature structure *satisfies* the equation

$$x_1 w_1 = x_2 w_2 \tag{1}$$

if and only if $\delta(f(x_1), w_1) = \delta(f(x_2), w_2)$, and the equation

$$x_3 w_3 = v \tag{2}$$

if and only if $\alpha(\delta(f(x_3), w_3)) = v$, where $x_1, x_2, x_3 \in D$, $w_1, w_2, w_3 \in A^*$ and $v \in V$. We only allow equations on those two forms, $x_1 w_1 = x_2 w_2$ and $x_3 w_3 = v$. In the grammar formalism we will even limit this a bit more.

If $E$ is a set of equations and $M$ is a well defined feature structure such that $M$ satisfies every equation in $E$ then we say that $M$ *supports* $E$ and we write

$$M \models E \tag{3}$$

If $M_1$ and $M_2$ are feature structures, we say that $M_1$ *subsumes* $M_2$, written $M_1 \sqsubseteq M_2$, if and only if for every set of equations $E$, if $M_1 \models E$ then $M_2 \models E$. If $M_1$ subsumes $M_2$ then $M_2$ contains all the information that $M_1$ contains. We see that if $M_1 \models E$ so must $M_2 \models E$ for all $M_2$ such that $M_1 \sqsubseteq M_2$. Two feature structures $M_1$ and $M_2$, are *equivalent* if and only if $M_1 \sqsubseteq M_2$ and $M_2 \sqsubseteq M_1$. This means that they contain the same information. Then subsumption will give us a partial order of the equivalent classes of feature structures. Given a set of equations $E$, we say that $E$ *describes* a feature structure $M$ if and only if $M \models E$ and for every feature structure $M'$, if $M' \models E$ then $M \sqsubseteq M'$. A given set of equations describes different, but equivalent feature structures. If a equation set is supported by an feature structure, then there exists a feature structure which the equation set describes. If $E$ describes a feature structure $M$ we write this

$$E \gg M \tag{4}$$

Here we describe feature structures without using unification. In our very simple way of defining and describing feature structures this is only a matter of taste and unification is just another approach to the same kind of information collecting. To see this, we define unification in the usual way: Let $M_1$ and $M_2$ be two well defined feature structures. Then the unification of $M_1$ and $M_2$, $(M_1 \sqcup M_2)$ is a feature structure such that $M_1 \sqsubseteq (M_1 \sqcup M_2)$, $M_2 \sqsubseteq (M_1 \sqcup M_2)$ and for every $M'$ such that $M_1 \sqsubseteq M'$ and $M_2 \sqsubseteq M'$, $(M_1 \sqcup M_2) \sqsubseteq M'$. From the definition we then get that if $E_1 \gg M_1$ and $E_2 \gg M_2$ then $(E_1 \cup E_2) \gg M_1 \sqcup M_2$. Instead of using unification directly we collect equations and see if all the equations together describe a feature structure.

A set of equations $E$ is *consistent* if there exists a well defined feature structure that $E$ describes. It is possible that an equation set does not describes any well defined feature structure. We then say that the equation set is *inconsistent*. This happens for instance if the equation set contains both the equations $ea = v$ and $eaa = v$ for a value symbol $v$. A feature structure that satisfies those two equations cannot be atomic.

# 3 The grammar formalism

The Tree Homomorphic Feature Structure Grammar[1] (THFSG) is based on LFG [KB82], but is much simplified. The main difference is that we have a strong restriction on the sets of equation schemata, we treat the lexical items in almost the same way as production rules, and we do not have the completeness and coherence constraints or anything like the functional uncertainty mechanism [KMZ87]. We have instead tried to make the formalism as simple as possible. This grammar formalism is very much like the grammar formalism $GF1$ defined by Colban [Col91] but there is one main difference; we accept empty right hand sides in the lexicon rules. This gives us the ability to describe a full abstract family of languages which $GF1$ does not [Bur92]. In LFG without functional uncertainty empty right hand sides are used in the analysis of long-distance dependencies. In addition $GF1$ only accepts equation schemata on the format that THFSG-grammars have in their normal form.

**Definition 2** *A Tree Homomorphic Feature Structure Grammar (THFSG) is a 5-tuple $\langle \mathcal{K}, \mathcal{S}, \Sigma, \mathcal{P}, \mathcal{L} \rangle$ over the set of attribute symbols $A$ and value symbols $V$ where*

- *$\mathcal{K}$ is a finite set of symbols, called* categories,
- *$\mathcal{S} \in \mathcal{K}$ is a symbol, called* start symbol,
- *$\Sigma$ is a finite set of symbols, called* terminals,
- *$\mathcal{P}$ is a finite set of* production rules

$$\begin{array}{rccc} K_0 & \rightarrow & K_1 & \ldots & K_m \\ & & E_1 & & E_m \end{array} \qquad (5)$$

*where $m \geq 1$, $K_0, ..., K_m \in \mathcal{K}$, and for all $i$, $1 \leq i \leq m$, $E_i$ is a finite set consisting of one and only one equation schema on the form $\uparrow a_1...a_n =\downarrow$ where $n \geq 0$ and $a_1, ..., a_n \in A$, and a finite number of equation schemata on the form $\uparrow a_1...a_n = v$ where $n \geq 1$, $a_1, ..., a_n \in A$ and $v \in V$.*

- *$\mathcal{L}$ is a finite set of* lexicon rules

$$\begin{array}{rcc} K & \rightarrow & t \\ & & E \end{array} \qquad (6)$$

*where $K \in \mathcal{K}$, $t \in (\Sigma \cup \{\varepsilon\})$, and $E$ is a finite set of equation schemata on the form $\uparrow a_1...a_n = v$ where $n \geq 1$, $a_1, ..., a_n \in A$ and $v \in V$.*

*The sets $\mathcal{K}$ and $\Sigma$ are required to be disjoint.*

As in LFG, we see that to each element on the right hand side in production and lexicon rules we annotate a set of equation schemata. These equation schemata differ form the equations used to describe feature structures: the schemata have up and down arrows where equations have names. The up and down arrows are metavariables: to get equations we instantiate the arrows to the nodes in the phrase structure tree. In the production rules each set of equation schemata includes one and only one schema with both up and down arrows. In this schema we only allow attribute symbols on the left hand side, –none on the right hand side. As a result of this we will later see that the described feature structure will be a tree that is homomorphic with the phrase structure tree or constituent structure as we will call it. But first we must define constituent structures and the set of grammatical strings with respect to a grammar.

---
[1]This grammar formalisms is part of a hierarchy of grammar formalisms based on different equation formats and definitions of grammatical strings described in [Bur92]. What we here call THFSG is there named $RS_1 \& T_0$. Among the other formalisms is $RS_0 \& T_0$, which has an undecidable membership question, and $RS_1 \& T_2$ for which we can decide membership in time $\mathcal{O}(n^3)$ and which in fact describes the class of context-free languages.

To define the constituent structures we use tree domains: Let $\mathcal{N}_+$ be the set of all integers greater than zero. A *tree domain* $D$ is a set $D \subseteq \mathcal{N}_+^*$ of number strings so that if $x \in D$ then all prefixes of $x$ are also in $D$, and for all $i \in \mathcal{N}_+$ and $x \in \mathcal{N}_+^*$, if $xi \in D$ then $xj \in D$ for all $j, 1 \leq j < i$. The *out degree* $d(x)$ of an element $x$ in a tree domain $D$ is the cardinality of the set $\{i \mid xi \in D, i \in \mathcal{N}_+\}$. The set of terminals of $D$ is $term(D) = \{x \mid x \in D, d(x) = 0\}$. The elements of a tree domain are totally ordered lexicographically as follows: $x' \prec x$ if $x'$ is a prefix of $x$, or there exist strings $y, z, z' \in \mathcal{N}_+^*$ and $i, j \in \mathcal{N}_+$ with $i < j$, such that $x' = yiz'$ and $x = yjz$.

A tree domain $D$ can be viewed as a tree graph in the following way: The elements of D are the nodes, $\varepsilon$ is the root, and for every $x \in D$ the element $xi \in D$ is $x$'s child number $i$. The terminals of $D$ are then the terminal nodes in the tree.

A tree domain describes the topology of a phrase structure tree. This representation provides a name for every node in the tree, directly from the definition of a tree domain. We will substitute the arrows used in the equation schemata with these names. A tree domain may be infinite, but we restrict the attention to finite tree domains.[2]

**Definition 3** *A* constituent structure *(c-structure) based on a* THFSG*-grammar* $G = \langle \mathcal{K}, \mathcal{S}, \Sigma, \mathcal{P}, \mathcal{L} \rangle$ *is a triple* $\langle D, K, E \rangle$ *where*

- *$D$ is a finite tree domain,*

- *$K : D \to (\mathcal{K} \cup \Sigma \cup \{\varepsilon\})$ is a function,*

- *$E : (D - \{\varepsilon\}) \to \Gamma$ is a function where $\Gamma$ is the set of all sets of equation schemata in $G$,*

*such that $K(x) \in (\Sigma \cup \{\varepsilon\})$ for all $x \in term(D)$, $K(\varepsilon) = \mathcal{S}$, and for all $x \in (D - term(D))$, if $d(x) = m$ then*

$$K(x) \to \begin{array}{ccc} K(x1) & ... & K(xm) \\ E(x1) & & E(xm) \end{array} \qquad (7)$$

*is a production or lexicon rule in $G$.*

*The terminal string of a constituent structure is the string $K(x_1)...K(x_n)$ such that $\{x_1, ..., x_n\} = term(D)$ and $x_i \prec x_{i+1}$ for all $i$, $1 \leq i < n$.*

Here the function $K$ labels the nonterminal nodes with category symbols and the terminal nodes with terminal symbols. The terminal string is then a string in $\Sigma^*$ since $K(x) \in (\Sigma \cup \{\varepsilon\})$ for all $x \in term(D)$. The function $E$ assigns a set of equation schemata to each node in the tree domain. This is done such that each mother-node together with all its daughters corresponds to a production or lexicon rule. To get equations that can be used to describe feature structures we must instantiate the up and down arrows in the equation schemata from the production and lexicon rules. We substitute them with nodes from the c-structure. For this purpose we define the '-function such that $E'(xi) = E(xi)[x/\uparrow, xi/\downarrow]$. We see that the value of the function $E'$ is a set of equations that feature structures may support.

---
[2]See Gallier [Gal86] for more about tree domains.

**Definition 4** *The c-structure $\langle D, K, E \rangle$ generates the feature structure $M$ if and only if*

$$\bigcup_{x \in (D - \{\varepsilon\})} E'(x) \gg M \tag{8}$$

*A c-structure is* consistent *if it generates a feature structure.*

The nonterminal part of the tree domain will form the name set for feature structures that this union describes. A c-structure is consistent if this union is consistent and a string is grammatical if its c-structure is consistent.

**Definition 5** *Let $G$ be a* THFSG-*grammar. A string $w$ is* grammatical *with respect to $G$ if and only if there exists a consistent c-structure with $w$ as the terminal string.*

The set of all grammatical strings[3] with respect to a grammar $G$ is denoted $L(G)$ and is the language that the grammar $G$ generates. Two grammars $G$ and $G'$ are *equivalent* if $L(G) = L(G')$.

**Example 1** Assume that *next* and *lex* are attribute symbols in $A$, and $a, b, c$ and \$ are value symbols in $V$. Let $G_1$ be the THFSG-grammar $\langle \mathcal{K}, S, \Sigma, \mathcal{P}, \mathcal{L} \rangle$ where $\mathcal{K} = \{S, B, B', C, C', C''\}$, $\Sigma = \{a, b, c\}$ and $\mathcal{P}$ contains the following production rules

$$S \quad \rightarrow \quad \begin{array}{cccc} B & C & C & B \\ \uparrow=\downarrow & \uparrow=\downarrow & \uparrow=\downarrow & \uparrow=\downarrow \end{array} \tag{9}$$

$$C \quad \rightarrow \quad \begin{array}{cc} C & C \\ \uparrow next = \downarrow & \uparrow next = \downarrow \end{array} \qquad C \quad \rightarrow \quad \begin{array}{c} C' \\ \uparrow=\downarrow \\ \uparrow next = \$ \end{array} \tag{10}$$

$$B \quad \rightarrow \quad \begin{array}{cc} B' & B \\ \uparrow=\downarrow & \uparrow next = \downarrow \end{array} \qquad B \quad \rightarrow \quad \begin{array}{c} B' \\ \uparrow=\downarrow \\ \uparrow next = \$ \end{array} \tag{11}$$

Moreover contains $\mathcal{L}$ the following lexicon rules

$$\begin{array}{cccc} B' & \rightarrow & a & \qquad B' \rightarrow & b \\ & & \uparrow lex = a & & \uparrow lex = b \\ \\ C' & \rightarrow & c & \\ & & \emptyset & \end{array} \tag{12}$$

---
[3]We may have different definitions of which strings are grammatical. For grammars in normal form (see below) we may also require for a c-structure to (correctly) generate a feature structure that for any two nodes $x$ and $y$ in the c-structure, if $\delta(f(x), w) = f(y)$ then $f(x) = f(x')$ where $x'$ is the greatest common prefix of $x$ and $y$, or in other words, $x'$ is the closest common predecessor. If we add this constraint we get a grammar formalism that describes the class of context-free languages [Bur92, Col91].

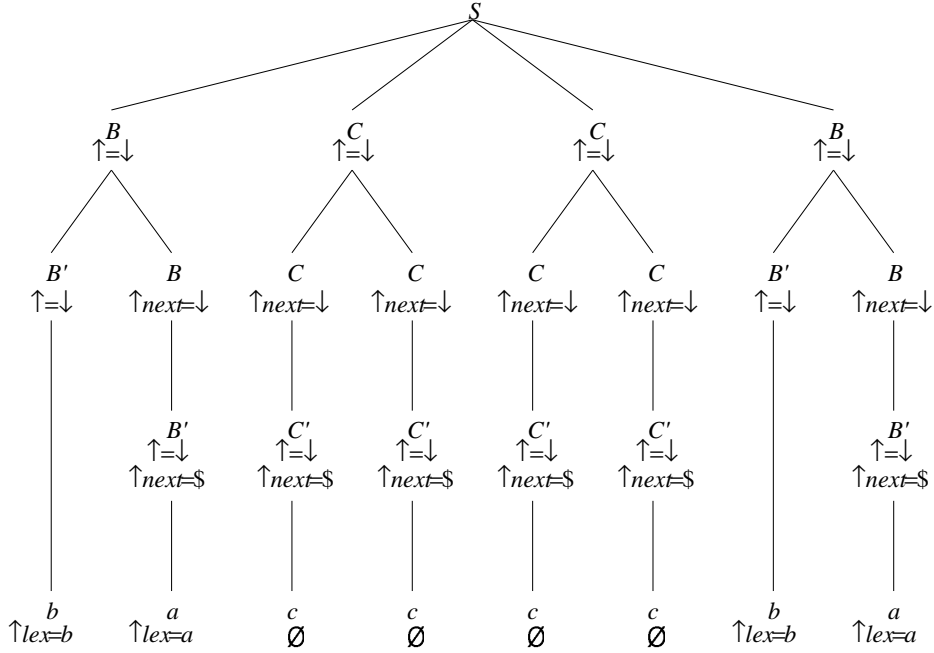

Figure 1: *c-structure for the string "baccccba" in $L(G_1)$.*

Figure 1 shows the c-structure for the string *baccccba*. The following are the equations we get from the left subtree after we have instantiated the up and down arrows:

$$\begin{aligned} \varepsilon &= 1 & 1\ next &= 12 \\ 1 &= 11 & 12 &= 121 \\ 11\ lex &= b & 12\ next &= \$ \\ & & 121\ lex &= a \end{aligned} \quad (13)$$

These are only a subset of all the equations from the c-structure. Figure 2 shows a feature structure which the c-structure generates. This shows that *baccccba* is grammatical with respect to $G_1$. The language generated by $G_1$ is

$$L(G_1) = \{wc^{2^n}w \mid w \in \{a,b\}^* \wedge |w| = n \wedge n \geq 1\} \quad (14)$$

Here we use the attribute *next* to count the length of the $w$ substring and the attribute *lex* to distribute information about its content.

In this grammar formalism we allow one and only one equation schema with both up and down arrows in each set of equation schemata in the production rules. Moreover in this schema we only allow attribute symbols on the left hand side — none on the right hand side. As a result the feature structures will be trees and the domination relation in the c-structure is preserved in the feature structure [Col91]. The domination relation must not be confused with the lexicographical ordering of the nodes in the c-structure, so let us define the domination relation on the c-structure and the feature structure: For all the nodes $x$ in the tree domain $D$ of a c-structure, let $x' \leq_c x$ for all prefixes $x'$ of $x$. This is the traditional predecessor relation on tree graphs. In the feature structure, let $q' \leq_M q$ for all nodes such that $\delta(q', w) = q$ for a $w \in A^*$. Then a node in a feature structure dominates another node if there exists an attribute path from the first node to the second node. For any c-structure $\langle D, K, E \rangle$ which generates a feature structure $M$ we then have

$$x' \leq_c x \;\Rightarrow\; f(x') \leq_M f(x) \tag{15}$$

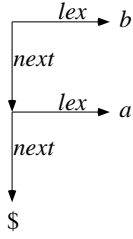

Figure 2: *Feature-structure for the string "baccccba" in $L(G_1)$. We have omitted the names here.*

Then the name function $f : D \to Q$ is a homomorphism between the node sets with the domination relation of those two structures [Col91].

We close this presentation of THFSG-grammars by defining a *normal form*:

**Definition 6** *A* THFSG-*grammar* $G = \langle \mathcal{K}, \mathcal{S}, \Sigma, \mathcal{P}, \mathcal{L} \rangle$ *is in normal form if each production rule in $\mathcal{P}$ is on the form*

$$\begin{array}{rcll} K_0 & \to & K_1 & K_2 \\ & & E_1 & E_2 \end{array} \tag{16}$$

*where $K_1, K_2, K_3 \in \mathcal{K}$, and each of the equation schema sets, $E_1$ and $E_2$, is a finite set consisting one and only one equation schema on the form $\uparrow a =\downarrow$ or $\uparrow=\downarrow$ where $a \in A$, and a finite number of equation schemata on the form $\uparrow a_1...a_n = v$ where $n \geq 1$, $a_1, ..., a_n \in A$ and $v \in V$.*

We see that a THFSG-grammar is in normal form if every production rule has exactly two elements on the right hand side and the equation schemata with both up and down arrows have no more than one attribute symbol.

**Lemma 1** *For every* THFSG-*grammar there exist an equivalent* THFSG-*grammar in normal form.*

**Proof:** We show how to construct a THFSG-grammar in normal form $G'$ for any THFSG-grammar $G$ such that $L(G) = L(G')$. There are two constraints for grammars in normal form, one on the equation schemata, and one on the format of the production rules. First we show how to get the equation schemata right.

For each set of equation schemata $E_i$ with an equation schema $\uparrow a_1...a_n =\downarrow$ where $n > 1$ in each production rule

$$\begin{array}{rcllll} K_0 & \to & K_1 & ... & K_i & ... & K_m \\ & & E_1 & & E_i & & E_m \end{array} \tag{17}$$

we replace $K_i$ with a unique new category $K'_{i,1}$ and $E_i$ with the set $E'_i = (E_i - \{\uparrow a_1...a_n =\downarrow\}) \cup \{\uparrow a_1 =\downarrow\}$, and add the new production rules:

$$K'_{i,(j-1)} \to \begin{array}{c} K'_{i,j} \\ \uparrow a_j =\downarrow \end{array} \qquad (18)$$

for all $j$, $2 \leq j \leq n$ where $K'_{i,2}...K'_{i,n-1}$ are unique new categories and $K'_{i,n} = K_i$. Now each set of equation schemata in each production rule is as required for the normal form.

Next, to get the production rules right: For each production rule

$$K_0 \to \begin{array}{ccc} K_1 & ... & K_m \\ E_1 & & E_m \end{array} \qquad (19)$$

with $m > 2$, we replace this production rule with the two production rules

$$K_0 \to \begin{array}{cc} K_1 & K'_2 \\ E_1 & \uparrow=\downarrow \end{array} \qquad (20)$$

$$K'_{m-1} \to \begin{array}{cc} K_{m-1} & K_m \\ E_{m-1} & E_m \end{array} \qquad (21)$$

together with the new production rules

$$K'_i \to \begin{array}{cc} K_i & K'_{i+1} \\ E_i & \uparrow=\downarrow \end{array} \qquad (22)$$

for all $i$, $2 \leq i \leq (m-2)$ where $K'_2, ..., K'_{m-1}$ are unique new categories. If $m = 1$ in the production rule (19) we replace the rule with the new production rule

$$K_0 \to \begin{array}{cc} K_1 & \tilde{\varepsilon} \\ E_1 & \uparrow=\downarrow \end{array} \qquad (23)$$

and add the lexicon rule

$$\tilde{\varepsilon} \to \begin{array}{c} \varepsilon \\ \emptyset \end{array} \qquad (24)$$

where $\tilde{\varepsilon}$ is a new category.

Now we have a grammar in normal form, and it is easy to see that we have a consistent c-structure for a string based on the original grammar if and only if we have a consistent c-structure for the same string based on the grammar in normal form. Then $L(G) = L(G')$.
□

## 4 Full abstract family of languages

When studying formal grammars we often want to study the class of languages that a grammar formalism defines. A *class of languages*, $\mathcal{C}_\Gamma$ over a countable set $\Gamma$ of symbols is a set of languages, such that for each language $L \in \mathcal{C}_\Gamma$ there exist a finite subset $\Sigma$ of $\Gamma$ such that $L \subseteq \Sigma^*$. The class $\mathcal{C}_\Gamma(GF)$ of languages that a grammar formalism $GF$ defines is the set of all languages $L'$ over $\Gamma$ such that there exists a grammar $G$ in $GF$ such that $L(G) = L'$.

For a given countably infinite $\Gamma$ an uncountable number of different classes of languages exist. Some of them are more natural and well-behaved than others, and of particular interest are the *full abstract families of languages* (full AFL). A full AFL is a class of languages closed under union, concatenation, Kleene closure, intersection with regular languages, string homomorphism and inverse string homomorphism[4]. The class of regular languages and context-free languages are both full AFL, but the class of context-sensitive languages is not since they are not closed under homomorphism[HU79]. Here we show that the class of languages that the grammar formalism THFSG defines[5], $\mathcal{C}(\text{THFSG})$ is a full abstract family of languages. But first we need a precise definition of full abstract families of languages.

A string homomorphism is a function $h : \Delta^* \to \Sigma^*$ such that for every $w \in \Delta^*$ and $a \in \Delta$ we have

$$h(\varepsilon) = \varepsilon \tag{25}$$
$$h(aw) = h(a)h(w) \tag{26}$$

A string homomorphic image of a language $L \subseteq \Delta^*$ for a string homomorphism $h : \Delta^* \to \Sigma^*$ is the language $\{h(w) \mid w \in L\}$. The inverse string homomorphic image of a language $L' \subseteq \Sigma^*$ is the language $\{w \mid h(w) \in L'\}$. The concatenation of two languages $L_1$ and $L_2$ is the language $\{w_1 w_2 \mid w_1 \in L_1 \wedge w_2 \in L_2\}$. The Kleene closure of a language $L$ is the language $\{w_1 \ldots w_n \mid n \geq 0 \wedge w_1, \ldots, w_n \in L\}$. Union and intersection are the standard set-theoretic operations.

**Lemma 2** $\mathcal{C}(\text{THFSG})$ *is closed under union, concatenation and Kleene-closure.*

**Proof:** Let two THFSG-grammars $G = \langle \mathcal{K}, \mathcal{S}, \Sigma, \mathcal{P}, \mathcal{L} \rangle$ and $G' = \langle \mathcal{K}', \mathcal{S}', \Sigma', \mathcal{P}', \mathcal{L}' \rangle$ be given and assume that $(\mathcal{K} \cap \mathcal{K}') = \emptyset$, $\mathcal{S}_0 \notin (\mathcal{K} \cup \mathcal{K}' \cup \Sigma \cup \Sigma')$, and that *first* and *next* are not used as attribute symbols in $G$ or $G'$.

*Union:* Let $G_\cup$ be the grammar $\langle \mathcal{K} \cup \mathcal{K}' \cup \{\mathcal{S}_0\}, \mathcal{S}_0, \Sigma \cup \Sigma', \mathcal{P}'', \mathcal{L} \cup \mathcal{L}' \rangle$ where $\mathcal{P}''$ is the least set such that $(\mathcal{P} \cup \mathcal{P}') \subseteq \mathcal{P}''$ and $\mathcal{P}''$ contains the following two production rules:

$$\mathcal{S}_0 \to \mathcal{S} \atop \uparrow = \downarrow \tag{27}$$

$$\mathcal{S}_0 \to \mathcal{S}' \atop \uparrow = \downarrow \tag{28}$$

Then $G_\cup$ is a THFSG-grammar and it is trivial that $L(G_\cup) = L(G) \cup L(G')$.

*Concatenation:* Let $G_{con}$ be the grammar $\langle \mathcal{K} \cup \mathcal{K}' \cup \{\mathcal{S}_0\}, \mathcal{S}_0, \Sigma \cup \Sigma', \mathcal{P}'', \mathcal{L} \cup \mathcal{L}' \rangle$ where $\mathcal{P}''$ is the least set such that $(\mathcal{P} \cup \mathcal{P}') \subseteq \mathcal{P}''$ and $\mathcal{P}''$ contains the following production rule:

$$\mathcal{S}_0 \to \mathcal{S} \quad \mathcal{S}' \atop \uparrow \textit{first} = \downarrow \quad \uparrow \textit{next} = \downarrow \tag{29}$$

Then $G_{con}$ is a THFSG-grammar and it is trivial that $L(G_{con}) = L(G)L(G')$.

*Kleene-closure:* Let $G_*$ be the grammar $\langle \mathcal{K} \cup \{\mathcal{S}_0\}, \mathcal{S}_0, \Sigma, \mathcal{P}'', \mathcal{L}'' \rangle$ where $\mathcal{P}''$ is the least set such that $\mathcal{P} \subseteq \mathcal{P}''$ and $\mathcal{P}''$ contains the following production rule:

$$\mathcal{S}_0 \to \mathcal{S} \quad \mathcal{S}_0 \atop \uparrow \textit{first} = \downarrow \quad \uparrow \textit{next} = \downarrow \tag{30}$$

---

[4] See Ginsburg [Gin75] for more about full abstract families of languages.
[5] We assume here that $\Gamma$ is the set of all symbols that we use and drop $\Gamma$ as subscript in $\mathcal{C}(\text{THFSG})$.

Moreover is $\mathcal{L}''$ the least set such that $\mathcal{L} \subseteq \mathcal{L}''$ and $\mathcal{L}''$ contains the following lexicon rule:

$$\mathcal{S}_0 \rightarrow \begin{array}{c} \varepsilon \\ \emptyset \end{array} \tag{31}$$

Then $G_*$ is a THFSG-grammar and it is trivial that $L(G_*) = L(G)^*$. □

To show that $\mathcal{C}(\text{THFSG})$ is closed under intersection with regular languages, string homomorphism and inverse string homomorphism we show that $\mathcal{C}(\text{THFSG})$ is closed under NFT-mapping. Informally, a *Nondeterministic Finite Transducer* (NFT) is a nondeterministic finite state machine with an additional write tape. In addition to just reading symbols and changing states, an NFT also writes symbols on the write tape. It may write symbols and change states when reading the empty string. Formally, an NFT is a 6-tuple $M = \langle Q, \Delta, \Sigma, \delta_0, q_0, F \rangle$ where $Q$ is a finite set of states, $\Delta$ is an input-alphabet, $\Sigma$ is an output-alphabet, $\delta_0$ is a function from $Q \times (\Delta \cup \{\varepsilon\})$ to finite subsets of $Q \times \Sigma^*$, $q_0 \in Q$ is the initial state and $F \subseteq Q$ is a set of final states.

For every $q_1, q_2, q_3 \in Q, a \in (\Delta \cup \{\varepsilon\}), w \in \Delta^*$ and $x, y \in \Sigma^*$, the extended transition function $\delta$ from $Q \times \Delta^*$ to subsets of $Q \times \Sigma^*$ is defined as the least function satisfying the following

$$(q_1, \varepsilon) \in \delta(q_1, \varepsilon) \tag{32}$$

$$(q_2, x) \in \delta(q_1, w) \land (q_3, y) \in \delta_0(q_2, a) \Rightarrow (q_3, xy) \in \delta(q_1, wa) \tag{33}$$

For any NFT $M = \langle Q, \Delta, \Sigma, \delta_0, q_0, F \rangle$, the NFT-mapping $M$ of a string $w \in \Delta^*$ and a language $L \subseteq \Delta^*$ is defined as follows:

$$M(w) = \{x \mid \exists q \in F : (q, x) \in \delta(q_0, w)\} \tag{34}$$

$$M(L) = \bigcup_{w \in L} M(w) \tag{35}$$

Further is the inverse NFT-mapping $M^{-1}$ of a string $x \in \Sigma^*$ and a language $L' \subseteq \Sigma^*$ defined as follows:

$$M^{-1}(x) = \{w \mid x \in M(w)\} \tag{36}$$

$$M^{-1}(L') = \bigcup_{w \in L'} M^{-1}(w) \tag{37}$$

The definition of NFT is sufficiently general that for any given NFT-mapping, the inverse NFT-mapping is also an NFT-mapping. A finite state machine is a special version of an NFT, which writes every symbol it reads, and does not change state or write anything while reading the empty string. If $M$ is a finite state machine version of an NFT, then $M(L)$ is the intersection of $L$ and the regular language that the finite state machine describes.

A string homomorphism $h : \Delta^* \rightarrow \Sigma^*$ can be expressed by an NFT. Let $M_h$ be the NFT $\langle Q, \Delta, \Sigma, \delta_0, q_0, F \rangle$ such that $Q = F = \{q_0\}$ and for all $a \in \Delta$ $\delta(q_0, a) = \{(q_0, h(a))\}$. Then $h(L) = M_h(L)$ for any language $L \subseteq \Delta^*$ and the inverse string homomorphism can also be expressed with an NFT-mapping.

By showing that the class $\mathcal{C}(\text{THFSG})$ is closed under NFT-mapping it follows that it is closed under intersection with regular languages, string homomorphism and inverse string homomorphism. We do this by first defining a grammar from a THFSG-grammar in normal form and an NFT, and then show that this grammar generates the NFT-mapping of the language generated by the first grammar.

**Definition 7** *Given a* THFSG*-grammar* $G = \langle \mathcal{K}, \mathcal{S}, \Delta, \mathcal{P}, \mathcal{L} \rangle$ *in normal form and a Nondeterministic Finite Transducer* $M = \langle Q, \Delta, \Sigma, \delta_0, q_0, F \rangle$. *Assume that the symbols* $\mathcal{S}_0$ *and* $\tilde{a}$ *for all* $a \in (\Sigma \cup \{\varepsilon\})$ *are not used in* $G$. *The grammar* $G_M = \langle \mathcal{K}', \mathcal{S}_0, \Sigma, \mathcal{P}', \mathcal{L}' \rangle$ *for the NFT-image* $M(L(G))$ *is defined as follows:*

*Let* $\mathcal{K}'$ *be the set* $(Q \times (\mathcal{K} \cup \Delta \cup \{\varepsilon\}) \times Q) \cup \{\tilde{a} \mid a \in (\Sigma \cup \{\varepsilon\})\} \cup \{\mathcal{S}_0\}$ *and let* $\mathcal{P}'$ *and* $\mathcal{L}'$ *be the least sets such that:*

a) *For all* $q \in F$, *the following is a rule in* $\mathcal{P}'$:

$$\mathcal{S}_0 \;\rightarrow\; \begin{array}{c} (q_0, \mathcal{S}, q) \\ \uparrow = \downarrow \end{array} \tag{38}$$

b) *For all production rules*

$$K_0 \;\rightarrow\; \begin{array}{cc} K_1 & K_2 \\ E_1 & E_2 \end{array} \tag{39}$$

*in* $\mathcal{P}$ *and all* $q_1, q_2, q_3 \in Q$, *the following is a rule in* $\mathcal{P}'$:

$$(q_1, K_0, q_3) \;\rightarrow\; \begin{array}{cc} (q_1, K_1, q_2) & (q_2, K_2, q_3) \\ E_1 & E_2 \end{array} \tag{40}$$

c) *For all lexicon rules*

$$K \;\rightarrow\; \begin{array}{c} b \\ E \end{array} \tag{41}$$

*in* $\mathcal{L}$ *and all* $q_1, q_2 \in Q$, *the following is a rule in* $\mathcal{P}'$:

$$(q_1, K, q_2) \;\rightarrow\; \begin{array}{c} (q_1, b, q_2) \\ E \cup \{\uparrow = \downarrow\} \end{array} \tag{42}$$

d) *For all* $q_1, q_2, q_3 \in Q$ *and all* $b \in (\Delta \cup \{\varepsilon\})$, *the following are rules in* $P'$

$$(q_1, b, q_3) \;\rightarrow\; \begin{array}{cc} (q_1, b, q_2) & (q_2, \varepsilon, q_3) \\ \uparrow = \downarrow & \uparrow = \downarrow \end{array} \tag{43}$$

$$(q_1, b, q_3) \;\rightarrow\; \begin{array}{cc} (q_1, \varepsilon, q_2) & (q_2, b, q_3) \\ \uparrow = \downarrow & \uparrow = \downarrow \end{array} \tag{44}$$

e) *For all* $q_1, q_2 \in Q, b \in (\Delta \cup \{\varepsilon\})$ *and* $y \in \Sigma^*$, *such that* $(q_2, y) \in \delta_0(q_1, b)$ *where* $y = a_1...a_n$ *for* $|y| = n \geq 1$, *or if* $y = \varepsilon$ *let* $\tilde{a}_1 = \tilde{\varepsilon}$ *and* $n = 1$, *the following is a production rule in* $\mathcal{P}'$:

$$(q_1, b, q_2) \;\rightarrow\; \begin{array}{ccc} \tilde{a}_1 & \cdots & \tilde{a}_n \\ \uparrow = \downarrow & & \uparrow = \downarrow \end{array} \tag{45}$$

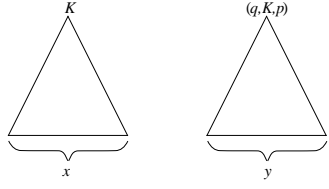

Figure 3: *Transformation to the NFT-mappings grammar for* $(p, y) \in \delta(q, x)$.

f) For all $a \in (\Sigma \cup \{\varepsilon\})$, the following is a rule in $\mathcal{L}'$:

$$\tilde{a} \rightarrow \begin{array}{c} a \\ \emptyset \end{array} \tag{46}$$

The main idea in this definition is that if a node in a c-structure based on $G$ with category $K$ is the root of a sub-c-structure with $x$ as terminal string and the NFT accepts $x$ as input string in a state $q$, then there is a corresponding node in a c-structure based on $G_M$ with category $(q, K, p)$. This node is the root of a sub-c-structure with $y$ as terminal string such that $(p, y) \in \delta(q, x)$, or less formally, such that the NFT may write $y$ when reading the string $x$ processing from state $q$ to $p$ (Figure 3). This is done such that the new c-structure gives a specification of how the NFT processes the input string, changes states and writes symbols. Downwards in the new c-structure we get more and more details of how the string is processed. In the end the grandmothers of the terminal nodes correspond to each transition step.

In the definition, part *a)*, *b)* and *c)* give us for any c-structure based on $G$ with $w = b_1...b_n$ as terminal string, the upper part of a new c-structure based on $G_M$, where the upper part is isomorphic with the first c-structure except that it will have an additional root node on the top. The main point here is that the terminal nodes in the first c-structure will have corresponding nodes with possible categories $(q_0, b_1, q_1), (q_1, b_2, q_2), ..., (q_{n-1}, b_n, q_n)$ in the new one, for any sequence of states, $q_0, q_1, ..., q_n$ where $q_0$ is the initial state, and $q_n$ is a final state. This is done such that if a node has (exactly) two daughters labeled $(q, K_1, q')$ and $(q'', K_2, q''')$, $q'$ must be equal to $q''$ and the mother node must be labeled $(q, K_0, q''')$ where $K_0, K_1$ and $K_3$ are the categories labeling the corresponding nodes in the first c-structure. Part *d)* in the definition allows the NFT to write symbols and change states while reading the empty string. In part *e)* we limit the previous parts of the definition such that all c-structures must correspond to the transition function in the NFT. This is achieved by requiring that for any symbol $b \in (\Delta \cup \{\varepsilon\})$, the triple category $(q_1, b, q_2)$ can only label the grandmother nodes of the terminal nodes in a c-structure if in fact there exists a one step transition from state $q_1$ to $q_2$ while reading $b$. The daughters of this node have nonterminal categories representing the output symbols of this one step transition. The last part of the definition *f)* is just the lexical complement of part *e)*.

With respect to the sets of equation schemata we take these with us to the new c-structure such that we get the same constraints on the new c-structures as on the c-structure based on the original grammar.

**Lemma 3** $\mathcal{C}(\text{THFSG})$ *is closed under NFT-mapping.*

**Proof:** Given Definition 7 we have to show that for all strings in $u \in \Sigma^*$, $u \in L(G_M)$ if and only if there exist a string $w$ in $L(G)$ and a a final state $q \in F$ such that $(q, u) \in \delta(q_0, w)$.

($\Rightarrow$) Assume that we have a consistent c-structure based on $G_M$ with $u$ as terminal string such that $u \in L(G_M)$. 1) By an induction on the height of the nodes we have from $d$), $e$) and $f$) in Definition 7 that if a node with category $(q, b, q')$ is a root of a sub-c-structure with $y$ as terminal string, where $b \in (\Delta \cup \{\varepsilon\})$, then $(q', y) \in \delta(q, b)$. 2) By an induction top down in the c-structure we have from $a$), $b$), $c$), and $d$) in Definition 7 that for any horizontal node-cut of nodes labeled with triple categories $(q_1, \beta_1, q'_1), ..., (q_n, \beta_n, q'_n)$ where $\beta_1, ..., \beta_n \in (\mathcal{K} \cup \Delta \cup \{\varepsilon\})$, that $q'_i = q_{i+1}$ for all $i$, $1 \leq i \leq (i-1)$, $q_1$ is the initial state and $q'_n$ is a final state. 3) There exists a sequence of the topmost nodes with triple categories where each $\beta_i$ is in $(\Delta \cup \{\varepsilon\})$ and each node has a mother node with a category $(q'_i, K_i, q_i)$ for $K_i \in \mathcal{K}$. This sequence forms a node cut and if $(q_0, b_1, q_1), (q_1, b_2, q_2), ..., (q_{n-1}, b_n, q_n)$ are the categories labeling these nodes in lexicographical order, this sequence give us a string $w = b_1...b_n$ in $\Delta^*$. The concatenation of the terminal strings $y_1, ..., y_n$ of the sub-c-structures where these nodes are the roots is $u$. From the definition of the extended transition function and the induction in the first part we have that $(q_n, u) \in \delta(q_0, w)$. 4) By reversing Definition 7 $b$) and $c$) it is straightforward to construct a c-structure for $w$ based on $G$, and if the c-structure for $u$ generates a feature structure so must the one for $w$. Then $w \in L(G)$.

($\Leftarrow$) Assume that we have a $w \in L(G)$ and a final state $q$ such that $(q, u) \in \delta(q_0, w)$ for a string $u \in \Sigma^*$. Since $(q, u) \in \delta(q_0, w)$ there must be a processing of $w$ of the NFT with $u$ as output. Following the discussion of Definition 7 it is straightforward to construct a c-structure for $u$ based on $G_M$ which specifies the processing of $w$ in $M$. If the c-structure for $w$ generates a feature structure so must the new one also, since we do not add any substantial new equations. Then we have that $u \in L(G_M)$. $\square$

From Lemma 2 and Lemma 3 we have the main result in this section.

**Theorem 1** $\mathcal{C}(\text{THFSG})$ *is a full abstract family of languages.*

## 5 Cross-Serial Dependencies

During the last ten to fifteen years the discussion whether or not natural languages can be described by context-free grammars has been revived [GP82]. This discussion distinguishes between a grammars capacity to describe a language *strongly*, i.e., to describe the language as a structured set, or *weakly*, i.e., to describe the language as a set of strings. Cross-serial dependencies are one of the main characteristics used to show that context-free grammars are not capable of even weakly to describe natural language.

Cross-serial dependencies occur in languages like $\{xx \mid x \in \Sigma^*\}$[6] and $\{wa^m b^n x c^m d^n y \mid w, x, y, \in \Sigma^*, m, n \geq 1, a, b, c, d \in \Sigma\}$, but not in languages like $\{xx^R \mid x \in \Sigma^*\}$[7] where we have nested dependencies.

Shieber [Shi85] has shown that Swiss German has cross-serial dependencies on the syntax level, and therefore in a weak description of the language. This is due to two facts about Swiss German:

> "First, Swiss German uses case-marking (dative and accusative) on objects, just as standard German does; different verbs subcategorize for objects of different

---
[6] We assume that $\Sigma$ has more than one symbol
[7] If $x = a_1 \ldots a_n$ then $x^R = a_n \ldots a_1$.

*case. Second, Swiss German, like Dutch, allows cross-serial order for the structure of subordinate clauses. Of critical importance is the fact that Swiss German requires appropriate case-marking to hold even within the cross-serial construction.*" Shieber [Shi85] (page 334).

This occurs e.g. in the following subordinate clauses preceded by *"Jan säit das"* ("Jan says that"):[8]

| ...mer | em Hans | es huus | hälfed | aastriiche | |
|---|---|---|---|---|---|
| ...we | Hans(DAT) | the house(ACC) | helped | paint | (47) |

*...we helped Hans paint the house.*

Here the verb *hälfed* subcategorizes for an object in dative; *em Hans*, and the verb *aastriiche* subcategorizes for an object in accusative; *es huus*. Shieber shows that this dependency is robust and that it holds in quite complex clauses, as seen in this example:

| ...mer | d'chind | em Hans | es huus | | | | |
|---|---|---|---|---|---|---|---|
| ...we | the children(ACC) | Hans(DAT) | the house(ACC) | | | | |
| | | | | haend | wele | laa | hälfe | aastriiche |
| | | | | have | wanted | let | help | paint | (48) |

*...we have wanted to let the children help Hans paint the house.*

If we change the cases of the objects then the strings become ungrammatical. Shieber (p. 336) specifies 4 claims that this construction in Swiss German satisfies:

1. *"Swiss-German subordinate clauses can have a structure in which all the Vs follow all the NPs."*

2. *"Among such sentences, those with all dative NP's preceding all accusative NPs, and all dative-subcategorizing Vs preceding all accusative-subcategorizing Vs are acceptable."*

3. *"The number of Vs requiring dative objects (e.g.,* hälfe*) must equal the number of dative NPs (e.g.,* em Hans*) and similarly for accusatives (*laa *and* chind*)."*

4. *"An arbitrary number of Vs can occur in a subordinate clause of this type (subject, of course, to performance constraints)."*

Shieber then shows that any language that satisfies these claims cannot be context-free, since such languages allow constructions on the form $wa^m b^n x c^m d^n y$. Here we study the language $L$ which contains strings on the form

*Jan säit das mer $N_1 \ldots N_n$ es huus haend wele $V_1 \ldots V_n$ aastriiche* (49)

where $n \geq 1$ and $N_i \in \{$*em Hans, es Hans, d'chind*$\}$[9] and $V_i \in \{$*hälfe, laa*$\}$ for all $i$, $1 \geq i \geq n$, and such that $V_i =$*hälfe* if and only if $N_i =$*em Hans*.

We see that this is a subset of Swiss German with the right case marking and subcategorizing and that it satisfies Shiebers claims. Hence it cannot be context-free. To make it easier to study we use the following homomorphism[10]:

---
[8]All linguistic data are from Shieber [Shi85].
[9]For simplicity we define the constructions *em Hans, es Hans* and *d'chind* as atomic symbols.
[10]We can do this since our grammar formalism is closed under string homomorphism and inverse string homomorphism

$$\begin{aligned}
h(\textit{Jan säit das mer}) &= x \\
h(\textit{es huus haend wele}) &= y \\
h(\textit{aastriiche}) &= z \\
h(s) &= s \quad \text{for all } s \in (N_{all} \cup V_{all})
\end{aligned} \qquad (50)$$

where $N_{all}$ is the set $\{\textit{em Hans, es Hans, d'chind}\}$ and $V_{all}$ is the set $\{\textit{hälfe, laa}\}$. We then have that $h(L)$ is the following language:

$$\begin{aligned}
h(L) &= \{xN_1\ldots N_n y V_1\ldots V_n z \mid \\
& \quad n \geq 1 \wedge \\
& \quad \forall i\ 1 \leq i \leq n\ [N_i \in N_{all} \wedge V_i \in V_{all} \wedge (V_i = \textit{hälfe} \iff N_i = \textit{em Hans})]\}
\end{aligned} \qquad (51)$$

We construct the following THFSG-grammar $G = \langle \mathcal{K}, \mathcal{S}, \Sigma, \mathcal{P}, \mathcal{L}\rangle$ for the language $h(L)$: Let
$$\begin{aligned}
\Sigma &= \{\textit{em Hans, es Hans, d'chind, hälfe, laa}, x, y, z\}, \text{and} \\
\mathcal{K} &= \{\mathcal{S}, VP, V, NP, N, X, Y, Z\}
\end{aligned}$$
We have the following production rules in $\mathcal{P}$:

$$S \to \begin{array}{ccccc} X & NP & Y & VP & Z \\ \uparrow = \downarrow & \uparrow = \downarrow & \uparrow = \downarrow & \uparrow = \downarrow & \uparrow = \downarrow \end{array}$$

$$NP \to \begin{array}{cc} N & NP \\ \uparrow obj = \downarrow & \uparrow vcomp = \downarrow \end{array} \qquad NP \to \begin{array}{c} N \\ \uparrow obj = \downarrow \\ \uparrow vcomp = null \end{array}$$

$$VP \to \begin{array}{cc} V & VP \\ \uparrow obj = \downarrow & \uparrow vcomp = \downarrow \end{array} \qquad VP \to \begin{array}{c} V \\ \uparrow obj = \downarrow \\ \uparrow vcomp = null \end{array}$$

We have the following lexicon rules in $\mathcal{L}$:

$$N \to \begin{array}{c} \textit{em Hans} \\ \uparrow case = DAT \end{array} \qquad N \to \begin{array}{c} \textit{es Hans} \\ \uparrow case = ACC \end{array} \qquad N \to \begin{array}{c} \textit{d'chind} \\ \uparrow case = ACC \end{array}$$

$$V \to \begin{array}{c} \textit{la} \\ \uparrow obj\ case = ACC \end{array} \qquad V \to \begin{array}{c} \textit{hälfe} \\ \uparrow obj\ case = DAT \end{array}$$

$$X \to \begin{array}{c} x \\ \emptyset \end{array} \qquad Y \to \begin{array}{c} y \\ \emptyset \end{array} \qquad Z \to \begin{array}{c} z \\ \emptyset \end{array}$$

From this grammar we get that strings like *"x em Hans d'chind y hälfe laa z"* are grammatical, while a string like *"x es Hans d'chind y hälfe laa z"* is ungrammatical, because of an inconsistency in the equation set. In figure 4 we show the c-structure and feature structure for the string

$$\textit{"x d'chind em Hans y laa hälfe z"} \qquad (52)$$

This is not meant as an adequate linguistic analysis, but an example of how we may collect cross-serial information with a THFSG-grammar.

## 6 Summary and remarks

We have defined a grammar formalism that describes a full abstract family of languages and showed that it can weakly describe a small subset of Swiss German with cross-serial dependencies. The method used to show that THFSG describes a full abstract family of languages is of some independent interest. It seems to be applicable to many other unification grammar formalisms with a context free phrase structure backbone. The method basically requires that the equation sets are more or less uniform in the phrase structure and we have the possibility to add "no information" equation sets. Additional constraints on how information is collected, shared and distributed in the phrase-structure tree may complicate its application.

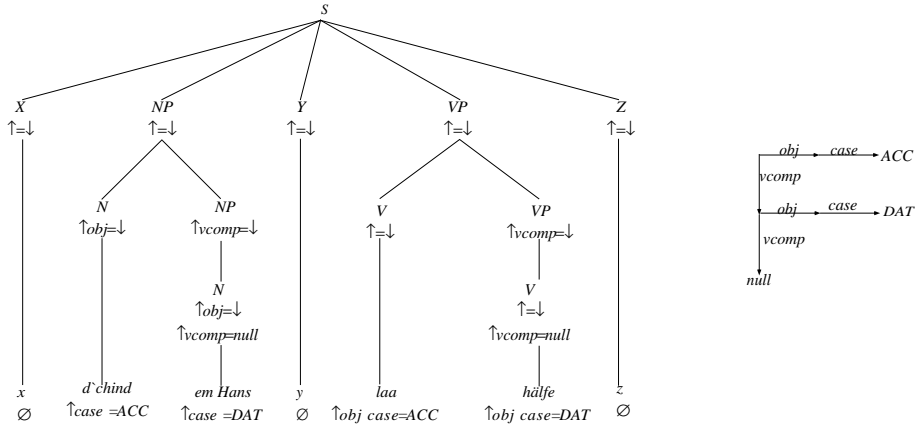

Figure 4: *c-structure and feature structure for cross-serial dependencies*

There are two potential disadvantages to THFSG. Firstly, its membership problem is NP-hard [Col91]. This due to the feature structures capacity to collect and distribute information across the sentence. This gives us the possibility to distribute truth-assignments uniformly for boolean expressions and then define a grammar that only accepts satisfiable expressions.

Secondly, does it have enough *linguistic flexibility*? By this we mean, is it possible in THFSG to express linguistic phenomena, as precisely as possible, in the way linguists would wish to state them? I THFSG we have a simple way of describing feature structures. As a result of this the feature structures will be trees. It may be argued that this is too limited compared to the much richer formalisms used in HPSG and LFG. On the other hand, on the string level of natural languages, cross-serial dependencies are to my knowledge the only constructions that are outside the context-free domain. Therefor the THFSG should be strong enough to describe the string sets of natural languages. However, we will not draw any strong conclusions regarding the linguistic adequacy of this grammar formalism, but leave it as an open question.

## Acknowledgments

I would like to thank Tore Langholm for his advice during the work that this paper is based on, and for extensive comments on earlier versions of the paper. I would also like to thank Patrick Blackburn. Without his interest this paper would still have been collecting dust on my shelf. He also made useful comments. The final preparation of this paper has been supported by grant 100437/410 from the Norwegian Research Council.